\documentclass[10pt]{article}
\usepackage[utf8]{inputenc}
\usepackage{graphicx}
\usepackage{indentfirst}
\usepackage{amsmath}
\usepackage{array}
\usepackage{multirow}
\usepackage{makecell}
\usepackage{float}
\usepackage{booktabs}
\usepackage{caption}
\usepackage{pdfpages}
\usepackage{enumerate}
\usepackage{svg}
\usepackage{authblk}
\usepackage{lipsum}

\newcommand{\keywords}[1]{\par\noindent\textbf{Keywords:} #1}
\usepackage[numbers,sort&compress]{natbib}
\usepackage{xcolor}
\usepackage[a4paper, margin=1.2in]{geometry}
\title{GaitASMS: Gait Recognition by Adaptive Structured Spatial Representation and Multi-Scale Temporal Aggregation}

\author[1*]{Yan Sun}
\author[1]{Hu Long}
\author[1]{Xueling Feng}
\author[2]{Mark Nixon}

\affil[1]{School of Computer Engineering and Science, Shanghai University, 99 Shangda Road, Shanghai, 200444, China}
\affil[2]{School of Electronics and Computer Science, University of Southampton, University Road, Southampton, SO17 1BJ, United Kingdom}

\date{%
    $^*$Corresponding author(s). E-mail(s): yansun@shu.edu.cn;\\%
    Contributing authors: longhu@shu.edu.cn; fengxueling@shu.edu.cn; msn@ecs.soton.ac.uk;\\[2ex]%
}

\begin{document}

\maketitle
\begin{abstract}
\noindent Gait recognition is one of the most promising video-based biometric technologies. The edge of silhouettes and motion are the most informative feature and previous studies have explored them separately and achieved notable results. However, due to occlusions and variations in viewing angles, their gait recognition performance is often affected by the predefined spatial segmentation strategy. Moreover, traditional temporal pooling usually neglects distinctive temporal information in gait. To address the aforementioned issues, we propose a novel gait recognition framework, denoted as GaitASMS, which can effectively extract the adaptive structured spatial representations and naturally aggregate the multi-scale temporal information. The Adaptive Structured Representation Extraction Module (ASRE) separates the edge of silhouettes by using the adaptive edge mask and maximizes the representation in semantic latent space. Moreover,  the Multi-Scale Temporal Aggregation Module (MSTA) achieves effective modeling of long-short-range temporal information by temporally aggregated structure. Furthermore, we propose a new data augmentation, denoted random mask, to enrich the sample space of long-term occlusion and enhance the generalization of the model. Extensive experiments conducted on two datasets demonstrate the competitive advantage of proposed method, especially in complex scenes, i.e. BG and CL. On the CASIA-B dataset, GaitASMS achieves the average accuracy of 93.5\% and outperforms the baseline on rank-1 accuracies by 3.4\% and 6.3\%, respectively, in BG and CL. The ablation experiments demonstrate the effectiveness of ASRE and MSTA. The source code is available at https://github.com/YanSun-github/GaitASMS.

\end{abstract}

\section{Introduction}
\noindent Gait recognition is one of the most popular biometric technologies, which can be employed for human identification at long-distance. Compared with other biometric modalities, \(e.g.\), fingerprint, face, and iris, gait does not require cooperation between target subjects and can be hard to disguise. With such advantages, gait recognition has enormous potential in access control, crime investigation, and social security. However, in real-world scenarios, variations like view changes, different wearing conditions, and occlusion \cite{ref1, ref2, ref3} can cause dramatic changes in gait silhouettes, which are significant challenges to gait recognition.
\par Nowadays, many deep learning-based gait recognition frameworks have been proposed to generate discriminative gait feature representations \cite{ref4, ref5, ref6, ref7}. \textbf{Spatial Feature Extraction:} \textbf{1)} There are some studies \cite{ref8, ref9, ref10} that take the frame-level features as a whole unit for feature extraction. Since the differences between gait silhouettes are tiny, simply using global feature extractors, \(e.g.\), global convolution layers and global max/mean pooling are ineffective in capturing fine-grained body information. \textbf{2)} The other studies \cite{ref5, ref6, ref7, ref11} are part-based gait recognition methods, which adopt a predefined segmentation strategy to partition the gait silhouettes for fine-grained gait representations. However, these methods only focus on specific body parts, which may limit the ability to capture the global gait features and neglect the relations among different parts. Moreover, since the silhouettes are easily affected by the wearing conditions and the change of viewpoint, it is difficult for the predefined segmentation strategy to effectively focus on the corresponding local features when the contours change dramatically. \textbf{Temporal Feature Extraction:} Most state-of-the-art gait recognition methods extract temporal features from gait sequences \cite{ref5, ref6, ref7, ref8, ref9, ref10, ref11}. Gait sequences contain rich temporal information, but these methods only use short-term temporal extractors or simple downsampling pooling operations to extract temporal gait representations. The argument regarding short-term temporal modeling can incorporate the premise that it does not focus on the periodic motion of gait, which is essential but often disrupted by occlusion. Consequently, the use of short-term temporal window functions exhibits a fundamental weakness, which makes these models exhibit poor robustness to occlusion.
\par To alleviate these issues, we propose a novel gait recognition framework, GaitASMS, which can extract the adaptive spatial representations from global and local features and aggregate long-short-range temporal features in the gait sequences. Specifically, the Adaptive Structured Representation Extraction Module (ASRE) is presented, which adopts an edge-based attention mechanism to extract local fine-grained gait features containing dynamic information and utilizes 3D convolutions for global gait representation extraction. Furthermore, we also propose a temporal aggregate module, called the Multi-Scale Temporal Aggregation Module (MSTA), to achieve long-short-range temporal information aggregation. The captured rich temporal information can effectively compensate for the missing silhouette information caused by occlusion. Transformer-based methods \cite{ref36} have become increasingly popular in the field of computer vision recently. TransGait \cite{ref34} employs the transformer module to fuse multi-modal visual information. Compared with the transformer architecture, MSTA is simple but efficient and it does not require massive computation resources for support. Additionally, a new data augmentation method is proposed, denoted as the random mask, which randomly masks some regions of input gait sequences in the target subject level to improve the robustness of the model to occlusion.

\par The main contributions of this paper can be summarized as the following four aspects:
\begin{enumerate}
    \item [\textbf{1)}]
    \textbf{In ASRE,} the local feature extractor adopts an edge-based attention mechanism to obtain adaptive edge masks and extract fine-grained edge gait representations. The global feature extractor is utilized to extract global spatial information as supplementary information to local gait features.
    
    \item [\textbf{2)}]
    \textbf{In MSTA,} we argue that the short-range temporal feature includes subtle variations in gait and the long-range temporal information contains the robust gait representation for occlusion. Therefore, we introduce the Multi-Scale Temporal Aggregation Module (MSTA) to effectively aggregate the long-short-range temporal information for extracting discriminative temporal features and enhancing the robustness of the model to occlusion.

    \item [\textbf{3)}]
    \textbf{In random mask,} as a novel data augmentation method, it is introduced to enrich the sample space of long-term occlusion and enhance the generalization of the model. Different from traditional masking operations, the random mask exhibits randomness only at the level of target objects, while at the sequence level, the masking regions remain fixed. This design effectively simulates scenarios with long-term occlusion.

    \item [\textbf{4)}]
    \textbf{In GaitASMS,} we adopt adaptive structured spatial representation extraction and multi-scale temporal aggregation to extract distinctive gait features. Extensive experiments conducted on the widely used gait public datasets, CASIA-B and OU-MVLP, demonstrate that GaitASMS outperforms prior SOTA gait recognition methods, especially in occlusion conditions. Abundant ablation experiments conducted on CASIA-B further prove the superiority of the proposed modules.
\end{enumerate}

\section{Related Work}
\noindent \textbf{Gait Recognition.} According to the type of input data, current gait recognition methods can be classified into two types, \textit{i.e.}, model-based and appearance-based. \textbf{Model-based} methods \cite{ref12, ref13, ref14, ref15, ref16} take the structure of human body as input, like 2D/3D pose data. In theory, model-based gait recognition methods have higher natural robustness to occlusion due to the absence of appearance information. However, considering the difficulty of accurately extracting human pose information from low-resolution images and complex scenes, the practicality of the model-based methods is limited. \textbf{Appearance-based} methods \cite{ref5, ref6, ref7, ref11, ref17} utilize RGB images or gait silhouettes as input, and attempt to extract discriminative gait features from them. As appearance-based methods do not require extra pose information extraction, it is relatively low-cost and operationally convenient. However, these methods heavily rely on silhouette information, thus they are extremely sensitive to occlusion. Our proposed approach belongs to the appearance-based method and effectively addresses the above issues. 
\par Most state-of-the-art works have taken spatial feature extraction and temporal modeling as the focus. Below, we will provide a detailed introduction to these two forms of feature extraction. \textbf{In spatial feature extraction,} recently proposed gait recognition methods can be divided into two categories: global-based and local-based. The global-based methods \cite{ref17, ref18, ref19, ref20, ref21} extract gait features from the whole human body. Specifically, GaitSet \cite{ref17} utilizes 2D CNN to capture frame-level spatial features, and then aggregates spatial gait representations by multiple statistical functions. MT3D \cite{ref18} uses 3D CNN to obtain spatial-temporal gait features from whole gait sequences. GLN \cite{ref21} extracts the silhouette-level and set-level features in different stages, and then merges them by the lateral connections in a top-down manner. GaitEdge \cite{ref35} uses edge detection to pre-process the gait silhouette to capture its shape and texture information, which may be useful in shallow networks, but as the depth of the network increases, it is difficult to truly extract the most discriminatory gait representations without adaptively adjusting the edge segmentation region. The local-based methods \cite{ref5, ref22, ref23, ref24} usually adopt a predefined segmentation strategy to obtain the parts of human silhouettes and then extract fine-grained features from each part. For instance, GaitPart \cite{ref5} proposes a focal convolution layer, which is used for obtaining fine-grain gait representations. 3DLocal \cite{ref24} implements a simple but effective form of 3D local CNNs for capturing detailed gait features from multi-scale local parts. GaitStrip \cite{ref22} learns the local-based spatial features by directly taking each strip of the human body as the basic unit. However, the aforementioned methods all have some issues: \textbf{(1)} The global-based methods have difficulty capturing fine-grained spatial features of gait. And the issue becomes more prominent as the network gets deeper. \textbf{(2)} The local-based methods encounter challenges in capturing correlation information among local regions and adaptively extracting the most discriminative local features. Particularly, although GaitGL \cite{ref23} can extract both global and local spatial features, it struggles to adaptively adjust the segmentation strategy according to the degree of occlusion.
\par \textbf{In temporal modeling,} these approaches can be divided into three categories: GEI-based, Set-based, and Sequence-based. The GEI-based gait recognition methods \cite{ref19, ref25} obtain Gait Energy Image (GEI) by aggregating all temporal information of a sequence. Then, it extracts the final gait representations from the GEI. The set-based methods \cite{ref17, ref20} treat the entire gait sequence as an unordered set to extract the temporal features. Recently, sequence-based gait recognition methods have outperformed alternative approaches, leading to their emergence as the mainstream approach in gait recognition. GaitPart \cite{ref5} adopts Micro-motion Capture Module (MCM) to capture short-range temporal information from each sequence. GaitGL \cite{ref23} utilizes Local Temporal Aggregation (LTA) operation to aggregate local temporal information. However, the methods based on local temporal modeling often perform poorly under occlusion conditions because they have difficulty in effectively learning long-term temporal correlation information to compensate for the missing information caused by occlusion.

\par To address these issues, we propose a novel Adaptive Structured Representation Extraction Module (ASRE) to capture the most dynamic gait patterns and generate adaptive structured spatial representations. And inspired by MG-TCN \cite{ref26}, we first introduce the Multi-Scale Temporal Aggregation Module (MSTA) from Re-ID to gait recognition and effectively achieve the long-short-range temporal modeling. Furthermore, an effective data augmentation approach is proposed to enlarge the sample of occlusion data and enable each module to be trained fully.

\begin{figure*}[bpht]
    \centering
    \includegraphics[width=1\textwidth]{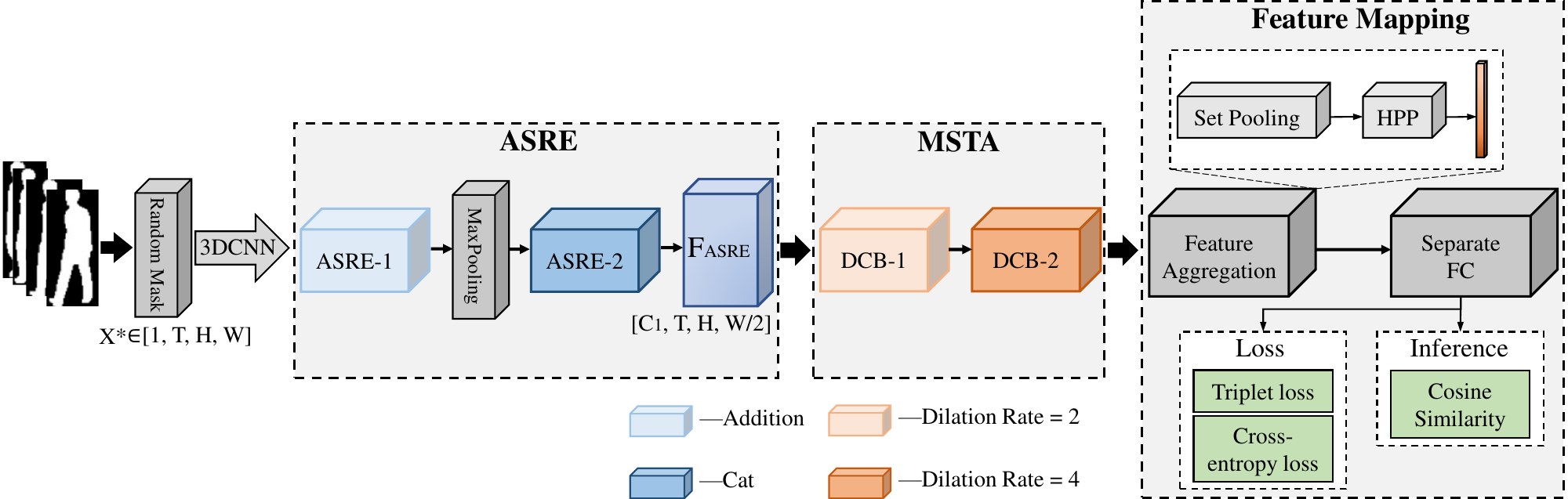}
    \caption{The overview of our GaitASMS. "ASRE" represents the Adaptive Structured Representation Extraction Module. "Addition" means the element-wise addition of local features and global features. "Cat" indicates combining local features and global features in the H dimension. "MSTA" represents the Multi-Scale Temporal Feature Aggregation Module, which is composed of the dilated convolution residual blocks. "HPP" means horizontal pyramid pooling \cite{ref27}. }
    \label{fig:pipeline}
\end{figure*}

\section{Proposed Method}
\noindent In this section, we first introduce the pipeline of the proposed method, which can be used to generate discriminative spatial gait representation and extract long-short range temporal correlations. Then the key modules are described, including the Adaptive Structured Representation Extraction Module (ASRE) and the Multi-Scale Temporal Aggregation Module (MSTA). Moreover, a new data augmentation is also proposed. Finally, the details of feature mapping, training, and testing are explained.

\subsection{Pipeline}
\noindent The pipeline of GaitASMS is shown in Fig.\ref{fig:pipeline}. GaitASMS is composed of two modules, including ASRE and MSTA. Firstly, the sequences of gait silhouettes are fed into the random mask for data augmentation and ASRE extracts adaptive structured spatial representation from the preprocessed data. Max Pooling is added to the framework for obtaining high-level spatial features and reducing the computational cost. Then, to achieve long-short temporal modeling, MSTA extracts the multi-scale temporal features from high-level spatial feature maps. In the end, feature aggregation and separate FC layers are used to map the feature vectors to embedding space for the final individual identification.

\par Assume that the input of GaitASMS is \(X=\left\{X_{1}, \ldots, X_{T}\right\}\), \(X \in R^{1 \times T \times H \times W}\), where \(T\) is the length of the sequence, \(H\) and \(W\) are the height and width of each frame, respectively. The pipeline of GaitASMS can be represented as
\begin{equation}
    X^{\ast } =Random Mask\left ( X \right )
\end{equation}
\begin{equation}
    F_{A S R E}=A S R E_{2}\left({MaxPooling}\left(A S R E_{1}(X^{\ast})\right)\right)
\end{equation}
\begin{equation}
    F_{M S T A}=D C B_{2}\left(D C B_{1}\left(F_{A S R E}\right)\right)
\end{equation}
\begin{equation}
    F_{gait}=F C\left(F A\left(F_{M S T A}\right)\right)
\end{equation}
where \(RandomMask\left(\cdot\right)\) applies masks of the same size to different positions of randomly selected subjects for masking operations. The difference between \(A S R E_{1}\left ( \cdot \right )\) and \(A S R E_{2}\left ( \cdot \right )\) is the fusion way of local features and global features. \(D C B_{1} \left ( \cdot \right ) \) and \(D C B_{2} \left ( \cdot \right )\) represent the dilated convolution blocks, whose dilation rates are 2 and 4, respectively. \(F A \left ( \cdot \right ) \) and \(F C \left ( \cdot \right ) \) are feature aggregation and feature mapping, respectively. \(F_{gait}\) represents the final gait representation.

\subsection{Adaptive Structured Spatial Extraction Module}
\begin{figure}[htb]
    \centering
    \includegraphics[width=0.8\textwidth]{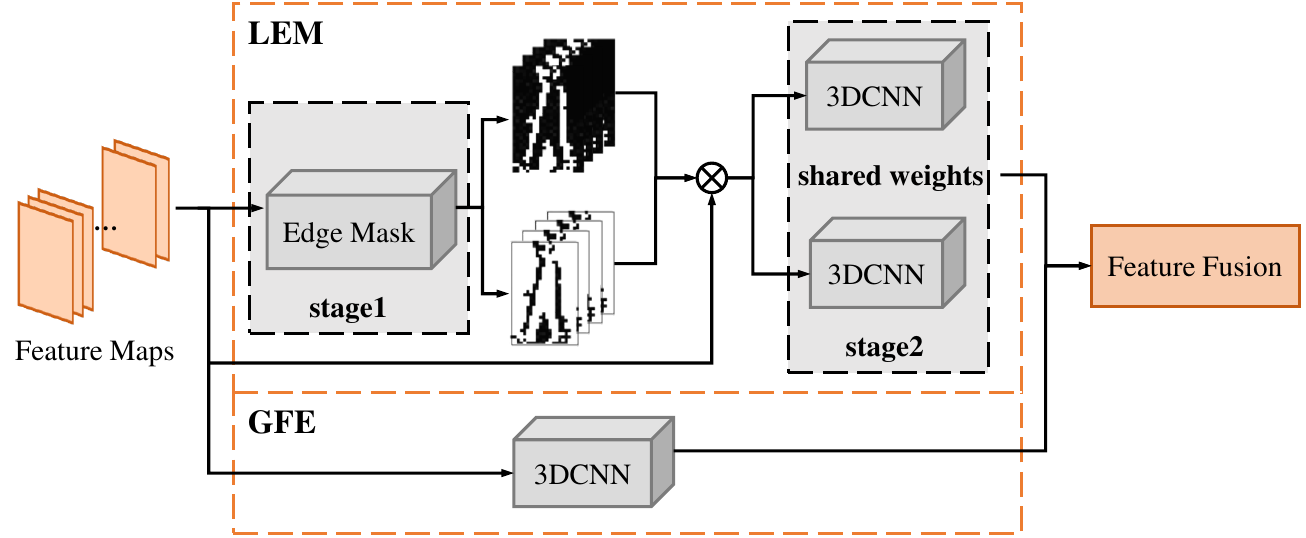}
    \caption{Overview of the ASRE. LEM is the Local Feature Extractor Based on Edge Mask. GFE is the Global Feature Extractor.} 
    \label{fig:ASRE}
\end{figure}

\noindent At present, most local-based gait recognition methods only use a predefined segmentation strategy to get the parts of human silhouettes. Although these methods can effectively extract fine-grained features from the parts, they cannot adaptively adjust the size and shape of the local area according to the occlusion status, which undoubtedly limits the representation ability of the network. Thus, the Adaptive Structured Representation Extraction Module (ASRE) is proposed for segmenting the edge of human contours and generating adaptive structured spatial representations. ASRE is shown in Fig.\ref{fig:pipeline}, which includes the Local Feature Extractor Based on Edge Mask (LEM) and the Global Feature Extractor (GFE). LEM can generate specific edge masks for different sequences, and then the edge masks are employed to obtain local features by masking the whole feature maps. The detailed structure is shown in Fig.\ref{fig:ASRE}. Thus, ASRE can effectively capture the most distinctive spatial features of each local part, even if under occlusion. GFE is used to extract frame-level spatial information as complementary information of local spatial features.

\par Inspired by GaitGL \cite{ref23}, we adopt a similar concatenation strategy as GLConv to fuse local spatial features and global spatial features in ASRE. However, the proposed ASRE is largely different from GLFE within GaitGL. Unlike the local feature extractor in GaitGL, which is used to extract details from predefined local maps, the LEM within ASRE generates adaptive structured spatial fine-grained representations through the edge-based attention mechanism. In addition, the subsequent experimental results demonstrate that ASRE can effectively replace GLFE and achieve better performance.

\par Assume that the input feature map is \(X^{\ast}=\left\{X_{1}^{\ast}, \ldots, X_{T}^{\ast}\right\}\), \(X_{i}^{\ast} \in R^{C \times T \times H \times W}\). ASRE can be represented as
\begin{equation}
    F_{A S R}=\left\{\begin{array}{ll}
    a d d \left(F_{G F E}, F_{L E M}\right), & A S R E_{1} \\
    c a t\left(F_{G F E}, F_{L E M}\right), & A S R E_{2}
    \end{array}\right.
\end{equation}
where \(F_{G F E}\) and \(F_{L E M}\) are global feature representation and local feature representation, respectively. \(cat\left ( \cdot \right )\) represents the concatenation operation on the \(H\) dimension.

\par
\begin{figure}[htb]
    \centering
    \includegraphics[width=0.8\textwidth]{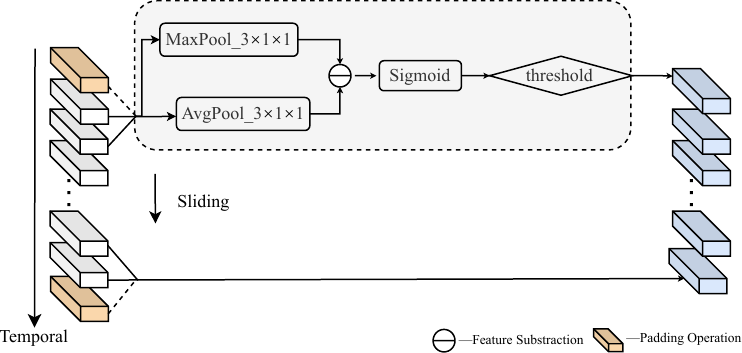}
    \caption{Operation of the Edge Mask.}
    \label{fig:EM}
\end{figure}

 As shown in Fig.\ref{fig:ASRE}, the LEM has two stages. \(stage_{1}\) generates the edge masks for each sequence. Specifically, \(MaxPool\left ( \cdot \right )\) and \(AvgPool\left ( \cdot \right )\) are used as aggregators to extract statistical information in the temporal dimension. Then using \(sigmoid\left ( \cdot \right )\) to normalize the difference between \(MaxPool\left ( \cdot \right )\) and \(AvgPool\left ( \cdot \right )\). The threshold is set as a hyper-parameter to control the intensity of attention on the edge mask. For obtaining richer local feature maps, we also adopt the edge segmentation strategy to generate a complementary mask for the edge mask. \(stage_{1}\) can be formulated as
\begin{equation}
    S=sigmoid\left(Max Pool_{3 \times 1 \times 1}(X^{\ast})- Avg Pool_{3 \times 1 \times 1}(X^{\ast})\right)
\end{equation}
\begin{equation}
    M_{edge}=\left\{\begin{array}{l}
    1, S \geq threshold \\
    0, S < threshold
    \end{array}\right.
\end{equation} 
\begin{equation}
    \bar{M}_{e d g e}=1-M_{e d g e}
\end{equation}
where \(\bar{M}_{edge}\) and \(M_{edge}\) is a pair of complementary masks. The generation of \(M_{edge}\) is shown in Fig.\ref{fig:EM}. In \(stage_{2}\), a shared weight 3D convolution is employed to extract fine-grained spatial features from the local feature maps based on edge mask. Thus, the \(F_{L E M}\) is shown as follows
\begin{equation}
    F_{L E M}=3DConv_{local}^{k\times k\times k}\left(X^{\ast} \otimes M_{\text {edge }}\right)+3DConv_{local}^{k\times k\times k} \left(X^{\ast} \otimes \bar{M}_{\text {edge }}\right)
\end{equation}
where \(3DConv_{local}^{k\times k\times k}\left ( \cdot \right )\) is the shared 3D convolution layer with kernel size \(k\times k\times k\). \(\otimes\) indicates the element-wise multiplication operation. As for the global gait feature \(F_{G F E}\), a similar mechanism is applied with 3D convolution kernels,
\begin{equation}
    F_{G F E}=3DConv_{global}^{k \times k \times k}\left(X^{\ast} \right)
\end{equation}
where \(3DConv_{global}^{k\times k\times k}\left ( \cdot \right )\) indicates 3D convolution operation with kernel size \(k\times k\times k\).

\subsection{Multi-Scale Temporal Aggregation Module}
\noindent Most of the previous gait recognition methods only use conventional convolution for temporal feature extraction. Due to the limited receptive field, it is difficult to extract long-range temporal relationships. Moreover, the adjacent frames of gait sequences are generally collected in less than 0.04 seconds \cite{ref28}. In other words, since the difference between adjacent frames is extremely small, it is difficult to learn discriminative temporal features only by temporal modeling of adjacent frames. Thus, we proposed a Multi-Scale Temporal Aggregation Module (MSTA), composed of multi-scale dilated convolution blocks with the residual connection. As shown in Fig.\ref{fig:pipeline}, the module can efficiently capture long and short-term temporal information. It enables the network to obtain supplemental information through other frames when the contour part is missing, which effectively improves the robustness of the network to occlusion.

\par
\begin{figure}[htb]
    \centering
    \includegraphics[width=0.7\textwidth]{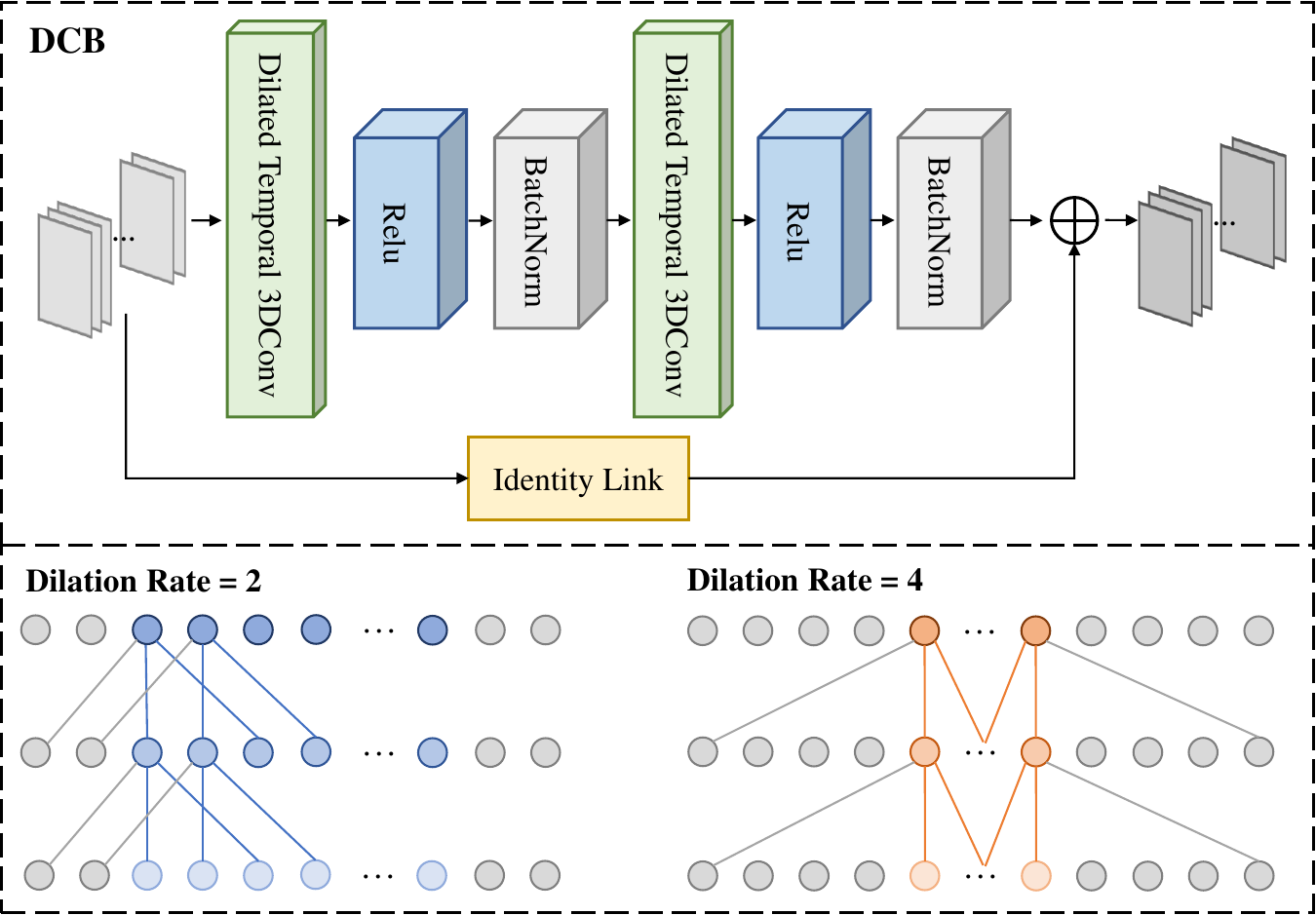}
    \caption{Overview of the DCB. It is the Dilated Convolution Block, which is composed of dilated 3D convolution layers, Relu, and BatchNorm.} 
    \label{fig:DCB}
\end{figure}

 The input of MSTA is \(F_{A S R E} \in R^{C_{1} \times T \times H \times W / 2}\), where \(C_{1}\) represents the channel, \((H,W/2)\) is the size of feature maps. \(F_{M S T A}\) can be expressed as
\begin{equation} 
    F_{M S T A} = D C B_{2}\left(D C B_{1}\left(F_{A S R E}\right)\right)
\end{equation}
where \(D C B_{1}\) and \(D C B_{2}\) are dilated convolution blocks, whose dilation rates are 2 and 4, respectively. The detailed pipeline of the \(D C B\) is depicted in Fig.\ref{fig:DCB}. In each \(D C B\), the input feature maps are processed sequentially through \(\left (Dilated\text{ }Temporal\text{ }3DConv, Relu, BatchNorm\right) * 2\), and the residual connection is also employed in \(D C B\).

\subsection{Gait Recognition Head}
\noindent A gait recognition head is employed to map the extracted feature maps into latent embedding space, generating the final gait representation \cite{ref28}. It generally includes temporal feature mapping and spatial feature mapping.
\par For the temporal feature mapping, we introduce a temporal Max-Pooling layer to aggregate the temporal information of variable-length gait sequences \cite{ref18}. Assuming that the feature map \(F_{M S T A} \in R^{C_{2} \times T \times H \times W / 2}\) is the final output of the feature extraction module. The temporal feature mapping is formulated as
\begin{equation}
    Y_{T F M}=MaxPool^{1 \times T \times 1 \times 1}\left(F_{M S T A}\right)
\end{equation}
where \(Y_{T F M} \in R^{C_{2} \times 1 \times H \times W / 2}\) is the output of the temporal feature mapping. \(MaxPool^{1 \times T \times 1 \times 1}\left ( \cdot  \right )\) performs a Max-Pooling operation on a sequence of length \(T\).
\par For the spatial mapping, we use a Generalized-Mean pooling (GeM) \cite{ref29} to generate multiple horizontal feature representations and then integrate the spatial information into the feature maps. Traditionally, most researchers fuse the features only by a weighted sum of \(MaxPool \left( \cdot \right )\) and \(AvgPool \left( \cdot \right )\). However, \(GeM\left( \cdot \right)\) can directly fuse these two different operations to form a feature map, with \(p = \infty \) being equal to \(MaxPool \left( \cdot \right )\) and \(p = 1\) being equal to \(AvgPool \left( \cdot \right )\),
\begin{equation}
    Y_{S F M}=F C\left(G e M\left(Y_{T F M}\right)\right)
\end{equation}
\begin{equation}
    G e M\left(Y_{T F M}\right)=\left ( AvgPool^{1 \times 1 \times 1 \times W/2}\left ( \left ( Y_{T F M} \right )^{p} \right ) \right ) ^{\frac{1}{p} } 
\end{equation}
where \(AvgPool^{1 \times 1 \times 1 \times W/2}\left ( \cdot \right ) \) is an average pooling operation with kernel size \(\left( 1 \times 1 \times 1 \times W/2 \right)\). \(F C \left ( \cdot \right )\) means the Fully Connected layer, which maps gait features into more discriminative embedding space for the final gait recognition.
\subsection{Training and Test Details}
\noindent \textbf{Training Loss.} In this paper, we introduce the combined loss function which consists of the triplet loss \cite{ref30} and cross-entropy loss to train the proposed network. The triplet loss is utilized to decrease the intra-class distances while increasing the inter-class distance, in addition to the cross-entropy loss employed for classification that facilitates the optimization of the model during training. The combined loss function is represented as
\begin{equation}
    L = L_{t r i} + L_{c s e}
\end{equation}
where \(L_{t r i}\) and \(L_{c s e}\) indicate the triplet loss and cross-entropy loss, separately. \(L_{t r i}\) is formulated as
\begin{equation}
    L_{t r i}=max \left\{d_{p}-d_{n}+margin, 0\right\}
\end{equation}
where \(d_{p}\) is the Euclidean distance between positive sample pairs, and \(d_{n}\) is the distance between negative sample pairs. The \(margin\) is a hyper-parameter that adjusts the optimization difficulty.
\par Similar to the implementation of OpenGait \cite{ref31}, we use Batch ALL as the sampling strategy. Each batch is formed as \((P\times K)\), where \(P\) is the number of subject classes and \(K\) denotes the number of samples for a subject class. Due to the memory limitation, the length of gait sequences is set to \(T\) during the training stage. The hardware platform consists of an Intel(R) Xeon(R) Silver 4110 CPU @ 2.10GHz, 64GB of RAM, and is equipped with 4 GTX 3090 Ti graphics cards.
\par \textbf{Test Stage.} In the test stage, we divide the gait datasets into gallery set and probe set for evaluation. It is worth noting that the samples from the gallery set are labeled data, while the samples in the probe set are used for prediction. The proposed network can extract the final gait representation from all frames of the samples belonging to different subject classes. Specifically, we calculate the Euclidean distance between the feature representations in the probe set and the other feature representations in the gallery set. The label of the gallery sample which has the smallest distance from this probe sample can be assigned to the probe sample. Finally, the Rank-1 recognition accuracy is calculated to evaluate the performance.

\section{Experiments}
\noindent In the section, two gait datasets, and the implementation details are first introduced. Then several ablation experiments are performed to show the effectiveness and robustness of GaitASMS.
\subsection{Datasets}
\par \noindent \textbf{CASIA-B.} Composed of 124 subjects, the CASIA-B \cite{ref32} is a widely applied gait dataset, in which each subject contains 11 views and each view contains 10 sequences. And the total 10 sequences are obtained under 3 walking conditions, \textit{i.e.}, normal walking (NM), walking with bags (BG), and walking with coats (CL). All video frames under each condition are captured by 11 fixed cameras and recorded with different viewpoints relative to the walking subject. For fairness, this paper strictly follows the popular protocol carried out by \cite{ref33}. In detail, the first 74 subjects are regarded as train sets, and the remaining 50 subjects are considered as test sets. During the testing phase, only the first four sequences in NM conditions are treated as gallery sets, and the others make up probe sets.

\par \noindent \textbf{OU-MVLP.} The OU-MVLP is one of the largest public gait datasets \cite{ref37}. However, similar to CASIA-B, it is also a cross-view dataset. It includes 10307 subjects (5153 subjects for training and the rest 5154 subjects for tests). In the dataset, each subject contains 14 views (\begin{math}0^{o}-90^{o}; 180^{o}-270^{o}\end{math}) and each view embodies 2 sequences (\#\textit{seq}-00, \#\textit{seq}-01)\cite{ref5}. In the testing phase, \#\textit{seq}-00 is regarded as gallery data.

\subsection{Implementation Details}

\par \noindent \textbf{Training details.} Common configuration: the gait silhouettes are normalized by \cite{ref17}, and resized to \(64\times44\). The same as \cite{ref28}, the \textit{margin} in Eqn.15 is set to 0.2 and the parameter \textit{p} in Eqn.13 is set to 6.5. Adam is taken as the optimizer and the length of each gait sequence T is set to 30. For the CASIA-B: the batch size \((P\times K)\) is \((8\times 8)\). The network structure on the CISIA-B is shown in Fig.\ref{fig:pipeline}. The output channels of "ASRE1", "ASRE2", "DCB1" and "DCB2" are 64, 128, 256, and 256, respectively. The learning rate (\(\lambda\)) is initialized to 1e-4 for the experiments and the total number of iterations is set to 80k. At 70k iterations, \(\lambda\) will be reduced to 1e-5. For the OU-MVLP: the batch size \((P\times K)\) is \((32\times 8)\). Because of the more subjects in OU-MVLP, we adopt the deeper network setting that doubles the number of the three modules i.e., "ASRE1", "ASRE2" and "DCB1". And the channels of the three modules are 64, 128, and 256. \(\lambda\) is initialized to 4e-4 and the total number of iterations is set to 130k. At 60k and 110k iterations, \(\lambda\) will be reduced to 4e-5 and 4e-6, respectively.

\subsection{Performance Comparison}
\noindent In this section, we compare our method with state-of-the-art ones on CASIA-B and OU-MVLP datasets, and the main results are given in Table \ref{tab:label_1} and Table \ref{tab:label_3}, respectively.

\textbf{Performances on CASIA-B}. We follow the dataset scales protocol \cite{ref23} of CASIA-B and test our method in all views and clothing conditions. The experimental results are shown in Table \ref{tab:label_1}. It can be observed that the proposed method has excellent performance with other SOTA methods in all clothing conditions. In the case of NM, BG and CL, our method outperforms the leading method GaitGL \cite{ref23} by 0.5\%, 1.3\%, and 3.1\%. Even compared with the bimodal gait recognition method TransGait \cite{ref34}, our method still achieves higher accuracy by 0.9\% and 0.9\% under the BG and CL conditions, respectively. The recognition accuracy of our method in these conditions achieves 97.9\%, 95.8\%, and 86.7\%, respectively. The results show that our method can effectively extract the discriminative gait features, and further improve the accuracy of the model in complex scenarios, specifically in the BG and CL. According to Table \ref{tab:label_2}, the average accuracy of our method is 93.5\%, which surpasses the advanced methods \text{i.e.}, GaitGL, and CSTL \cite{ref38} by 1.7\% and 1.6\%, respectively. Several contributing factors include: \textbf{1)} In complex scenarios, ASRE focuses on dynamic gait representations adaptively based on contextual information. \textbf{2)} Using the pyramid structure for temporal aggregation allows our model to capture richer temporal information and further improve its robustness to occlusions. \textbf{3)} The integration of random mask has extended the sample space of occlusion data, thereby boosting the generalization of our model.

\begin{table}[H]
    \centering
    \captionsetup{skip=2pt}
    \caption{Averaged Rank-1 accuracy (\%) on CASIA-B per probe angle excluding identical-view cases.}
    \label{tab:label_1}
    \resizebox{1\textwidth}{!}{
        \begin{tabular}{ll|llllllllllllp{0.4cm}}
            \toprule[1.5pt]
                \multicolumn{2}{c|}{Gallery}  & \multicolumn{11}{c|}{\begin{math}0^{o}-180^{o}\end{math}}  & \multirow{2}{*}{Mean} \\ \cline{1-13}
                \multicolumn{2}{c|}{Probe} & \multicolumn{1}{c|}{\begin{math}0^{o}\end{math}} & \multicolumn{1}{c|}{\begin{math}18^{o}\end{math}} & \multicolumn{1}{c|}{\begin{math}36^{o}\end{math}} & \multicolumn{1}{c|}{\begin{math}54^{o}\end{math}} & \multicolumn{1}{c|}{\begin{math}72^{o}\end{math}} & \multicolumn{1}{c|}{\begin{math}90^{o}\end{math}} & \multicolumn{1}{c|}{\begin{math}108^{o}\end{math}} & \multicolumn{1}{c|}{\begin{math}126^{o}\end{math}} & \multicolumn{1}{c|}{\begin{math}144^{o}\end{math}} & \multicolumn{1}{c|}{\begin{math}162^{o}\end{math}} & \multicolumn{1}{c|}{\begin{math}180^{o}\end{math}} \\ \hline\hline
                \multicolumn{1}{c|}{} & GaitSet\cite{ref20}   & \multicolumn{1}{c|}{90.8}  & \multicolumn{1}{c|}{97.9}   & \multicolumn{1}{c|}{99.4}   & \multicolumn{1}{c|}{96.9}   & \multicolumn{1}{c|}{93.6}   & \multicolumn{1}{c|}{91.7}   & \multicolumn{1}{c|}{95.0}    & \multicolumn{1}{c|}{97.8}    & \multicolumn{1}{c|}{98.9}    & \multicolumn{1}{c|}{96.8}    & \multicolumn{1}{c|}{85.8} & \multicolumn{1}{c}{95.0}    \\ \cline{2-14} 
                \multicolumn{1}{c|}{}  & GaitPart\cite{ref5}  & \multicolumn{1}{c|}{94.1}  & \multicolumn{1}{c|}{98.6}   & \multicolumn{1}{c|}{99.3}   & \multicolumn{1}{c|}{98.5}   & \multicolumn{1}{c|}{94.0}   & \multicolumn{1}{c|}{92.3}   & \multicolumn{1}{c|}{95.9}    & \multicolumn{1}{c|}{98.4}    & \multicolumn{1}{c|}{99.2}    & \multicolumn{1}{c|}{97.8}    & \multicolumn{1}{c|}{90.4} &   \multicolumn{1}{c}{96.2}   \\ \cline{2-14} 
                \multicolumn{1}{c|}{}  & MT3D\cite{ref18}     & \multicolumn{1}{c|}{95.7}  & \multicolumn{1}{c|}{98.2}   & \multicolumn{1}{c|}{99.0}   & \multicolumn{1}{c|}{97.5}   & \multicolumn{1}{c|}{95.1}   & \multicolumn{1}{c|}{93.9}   & \multicolumn{1}{c|}{96.1}    & \multicolumn{1}{c|}{98.6}    & \multicolumn{1}{c|}{99.2}    & \multicolumn{1}{c|}{98.2}    & \multicolumn{1}{c|}{92.0}     & \multicolumn{1}{c}{96.7}     \\ \cline{2-14} 
                \multicolumn{1}{c|}{}       & 3DLocal\cite{ref24}    & \multicolumn{1}{l|}{96.0}  & \multicolumn{1}{c|}{\underline{99.0}}   & \multicolumn{1}{c|}{\underline{99.5}}   & \multicolumn{1}{c|}{\underline{98.9}}   & \multicolumn{1}{c|}{\underline{97.1}}   & \multicolumn{1}{c|}{94.2}   & \multicolumn{1}{c|}{96.3}    & \multicolumn{1}{c|}{99.0}    & \multicolumn{1}{c|}{98.8}    & \multicolumn{1}{c|}{98.5}    & \multicolumn{1}{c|}{95.2}    & \multicolumn{1}{c}{97.5}     \\ \cline{2-14} 
                \multicolumn{1}{c|}{\textbf{NM}}     & GaitGL\cite{ref23}      & \multicolumn{1}{c|}{96.0}  & \multicolumn{1}{c|}{98.3}   & \multicolumn{1}{c|}{99.0}   & \multicolumn{1}{c|}{97.9}   & \multicolumn{1}{c|}{96.9}   & \multicolumn{1}{c|}{95.4}   & \multicolumn{1}{c|}{97.0}    & \multicolumn{1}{c|}{98.9}    & \multicolumn{1}{c|}{\underline{99.3}}    & \multicolumn{1}{c|}{98.8}    & \multicolumn{1}{c|}{94.0}    &   \multicolumn{1}{c}{97.4}   \\ \cline{2-14} 
                \multicolumn{1}{c|}{} & CSTL\cite{ref38} & \multicolumn{1}{c|}{\underline{97.2}}  & \multicolumn{1}{c|}{\underline{99.0}}   & \multicolumn{1}{c|}{99.2}   & \multicolumn{1}{c|}{98.1}   & \multicolumn{1}{c|}{96.2}   & \multicolumn{1}{c|}{\underline{95.5}}   & \multicolumn{1}{c|}{\underline{97.7}}    & \multicolumn{1}{c|}{98.7}    & \multicolumn{1}{c|}{99.2}    & \multicolumn{1}{c|}{98.9}    & \multicolumn{1}{c|}{\textbf{96.5}}    & \multicolumn{1}{c}{97.8}   \\ \cline{2-14} 
                \multicolumn{1}{c|}{} & TransGait\cite{ref34}   & \multicolumn{1}{c|}{\textbf{97.3}}  & \multicolumn{1}{c|}{\textbf{99.6}}   & \multicolumn{1}{c|}{\textbf{99.7}}   & \multicolumn{1}{c|}{\textbf{99.0}}   & \multicolumn{1}{c|}{\underline{97.1}}   & \multicolumn{1}{c|}{95.4}   & \multicolumn{1}{c|}{97.4}    & \multicolumn{1}{c|}{\underline{99.1}}    & \multicolumn{1}{c|}{\textbf{99.6}}    & \multicolumn{1}{c|}{98.9}    & \multicolumn{1}{c|}{95.8} & \multicolumn{1}{c}{\textbf{98.1}}    \\ \cline{2-14} 
                \multicolumn{1}{c|}{}  & \begin{math}\text{GaitStrip}^{*}\end{math}\cite{ref22}  & \multicolumn{1}{c|}{96.0}  & \multicolumn{1}{c|}{98.4}   & \multicolumn{1}{c|}{98.8}   & \multicolumn{1}{c|}{97.9}   & \multicolumn{1}{c|}{96.6}   & \multicolumn{1}{c|}{95.3}   & \multicolumn{1}{c|}{97.5}    & \multicolumn{1}{c|}{98.9}    & \multicolumn{1}{c|}{99.1}    & \multicolumn{1}{c|}{\underline{99.0}}    & \multicolumn{1}{c|}{\underline{96.3}}    &  \multicolumn{1}{c}{97.6}    \\ \cline{2-14} 
                \multicolumn{1}{c|}{} & \textbf{Ours}      & \multicolumn{1}{c|}{96.4}  & \multicolumn{1}{c|}{98.1}   & \multicolumn{1}{c|}{98.6}   & \multicolumn{1}{c|}{98.0}   & \multicolumn{1}{c|}{\textbf{97.3}}   & \multicolumn{1}{c|}{\textbf{96.7}}   & \multicolumn{1}{c|}{\textbf{98.6}}    & \multicolumn{1}{c|}{\textbf{99.4}}    & \multicolumn{1}{c|}{99.2}    & \multicolumn{1}{c|}{\textbf{99.6}}    & \multicolumn{1}{c|}{95.1}    & \multicolumn{1}{c}{\underline{97.9}}      \\ \hline\hline
                
                \multicolumn{1}{c|}{}  & GaitSet\cite{ref20}   & \multicolumn{1}{c|}{83.8}  & \multicolumn{1}{c|}{91.2}   & \multicolumn{1}{c|}{91.8}   & \multicolumn{1}{c|}{88.8}   & \multicolumn{1}{c|}{83.3}   & \multicolumn{1}{c|}{81.0}   & \multicolumn{1}{c|}{84.1}    & \multicolumn{1}{c|}{90.0}    & \multicolumn{1}{c|}{92.2}    & \multicolumn{1}{c|}{94.4}    & \multicolumn{1}{c|}{79.0}    &  \multicolumn{1}{c}{87.2} \\ \cline{2-14} 
                \multicolumn{1}{c|}{}   & GaitPart\cite{ref5}  & \multicolumn{1}{c|}{89.1}  & \multicolumn{1}{c|}{94.8}   & \multicolumn{1}{c|}{96.7}   & \multicolumn{1}{c|}{95.1}   & \multicolumn{1}{c|}{88.3}   & \multicolumn{1}{c|}{84.9}   & \multicolumn{1}{c|}{89.0}    & \multicolumn{1}{c|}{93.5}    & \multicolumn{1}{c|}{96.1}    & \multicolumn{1}{c|}{93.8}    & \multicolumn{1}{c|}{85.8}    &  \multicolumn{1}{c}{91.5} \\ \cline{2-14} 
                \multicolumn{1}{c|}{}  & MT3D\cite{ref18}  & \multicolumn{1}{c|}{91.0}  & \multicolumn{1}{c|}{95.4}   & \multicolumn{1}{c|}{\underline{97.5}}   & \multicolumn{1}{c|}{94.2}   & \multicolumn{1}{c|}{92.3}   & \multicolumn{1}{c|}{86.9}   & \multicolumn{1}{c|}{91.2}    & \multicolumn{1}{c|}{95.6}    & \multicolumn{1}{c|}{97.3}    & \multicolumn{1}{c|}{96.4}    & \multicolumn{1}{c|}{86.6}    & \multicolumn{1}{c}{93.0}\\ \cline{2-14} 
                \multicolumn{1}{c|}{}   & 3DLocal\cite{ref24}   & \multicolumn{1}{c|}{\underline{92.9}}  & \multicolumn{1}{c|}{95.9}   & \multicolumn{1}{c|}{\textbf{97.8}}   & \multicolumn{1}{c|}{96.2}   & \multicolumn{1}{c|}{93.0}   & \multicolumn{1}{c|}{87.8}   & \multicolumn{1}{c|}{92.7}    & \multicolumn{1}{c|}{96.3}    & \multicolumn{1}{c|}{97.9}    & \multicolumn{1}{c|}{\textbf{98.0}}    & \multicolumn{1}{c|}{88.5}    &   \multicolumn{1}{c}{94.3}   \\ \cline{2-14} 
                \multicolumn{1}{c|}{\textbf{BG}}   & GaitGL\cite{ref23}    & \multicolumn{1}{c|}{92.6}  & \multicolumn{1}{c|}{\underline{96.6}}   & \multicolumn{1}{c|}{96.8}   & \multicolumn{1}{c|}{95.5}   & \multicolumn{1}{c|}{93.5}   & \multicolumn{1}{c|}{89.3}   & \multicolumn{1}{c|}{92.2}    & \multicolumn{1}{c|}{96.5}    & \multicolumn{1}{c|}{\underline{98.2}}    & \multicolumn{1}{c|}{96.9}    & \multicolumn{1}{c|}{91.5}    &   \multicolumn{1}{c}{94.5}   \\ \cline{2-14} 
                \multicolumn{1}{c|}{}  & CSTL\cite{ref28}      & \multicolumn{1}{c|}{91.7}  & \multicolumn{1}{c|}{96.5}   & \multicolumn{1}{c|}{97.0}   & \multicolumn{1}{c|}{95.4}   & \multicolumn{1}{c|}{90.9}   & \multicolumn{1}{c|}{88.0}   & \multicolumn{1}{c|}{91.5}    & \multicolumn{1}{c|}{95.8}    & \multicolumn{1}{c|}{97.0}    & \multicolumn{1}{c|}{95.5}    & \multicolumn{1}{c|}{90.3}    &  \multicolumn{1}{c}{93.6}   \\ \cline{2-14} 
                \multicolumn{1}{c|}{} & TransGait\cite{ref34}   & \multicolumn{1}{c|}{\textbf{94.0}}  & \multicolumn{1}{c|}{\textbf{97.1}}   & \multicolumn{1}{c|}{96.5}   & \multicolumn{1}{c|}{96.0}   & \multicolumn{1}{c|}{93.5}   & \multicolumn{1}{c|}{\underline{91.5}}   & \multicolumn{1}{c|}{\underline{93.6}}    & \multicolumn{1}{c|}{95.9}    & \multicolumn{1}{c|}{97.2}    & \multicolumn{1}{c|}{97.1}    & \multicolumn{1}{c|}{\underline{91.6}} & \multicolumn{1}{c}{94.9}    \\ \cline{2-14} 
                \multicolumn{1}{c|}{}  & \begin{math}\text{GaitStrip}^{*}\end{math}\cite{ref22} & \multicolumn{1}{c|}{92.8}  & \multicolumn{1}{c|}{\underline{96.6}}   & \multicolumn{1}{c|}{97.2}   & \multicolumn{1}{c|}{\textbf{96.5}}   & \multicolumn{1}{c|}{\underline{95.2}}   & \multicolumn{1}{c|}{90.5}   & \multicolumn{1}{c|}{93.5}    & \multicolumn{1}{c|}{\underline{97.5}}    & \multicolumn{1}{c|}{\textbf{98.3}}    & \multicolumn{1}{c|}{97.6}    & \multicolumn{1}{c|}{91.4}    &  \multicolumn{1}{c}{\underline{95.2}}    \\ \cline{2-14} 
                \multicolumn{1}{c|}{} & \textbf{Ours}      & \multicolumn{1}{c|}{92.7}  & \multicolumn{1}{c|}{96.2}   & \multicolumn{1}{c|}{97.3}   & \multicolumn{1}{c|}{\underline{96.4}}   & \multicolumn{1}{c|}{\textbf{95.9}}   & \multicolumn{1}{c|}{\textbf{93.4}}   & \multicolumn{1}{c|}{\textbf{95.6}}    & \multicolumn{1}{c|}{\textbf{98.1}}    & \multicolumn{1}{c|}{\textbf{98.3}}    & \multicolumn{1}{c|}{\underline{97.7}}    & \multicolumn{1}{c|}{\textbf{91.7}}    & \multicolumn{1}{c}{\textbf{95.8}}    \\ \hline\hline
                
                \multicolumn{1}{c|}{} & GaitSet\cite{ref20}   & \multicolumn{1}{c|}{61.4}  & \multicolumn{1}{c|}{75.4}   & \multicolumn{1}{c|}{80.7}   & \multicolumn{1}{c|}{77.3}   & \multicolumn{1}{c|}{72.1}   & \multicolumn{1}{c|}{70.1}   & \multicolumn{1}{c|}{71.5}    & \multicolumn{1}{c|}{73.5}    & \multicolumn{1}{c|}{73.5}    & \multicolumn{1}{c|}{68.4}    & \multicolumn{1}{c|}{50.0}    &  \multicolumn{1}{c}{70.4}    \\ \cline{2-14} 
                \multicolumn{1}{c|}{}  & GaitPart\cite{ref5}  & \multicolumn{1}{c|}{70.7}  & \multicolumn{1}{c|}{85.5}   & \multicolumn{1}{c|}{86.9}   & \multicolumn{1}{c|}{83.3}   & \multicolumn{1}{c|}{77.1}   & \multicolumn{1}{c|}{72.5}   & \multicolumn{1}{c|}{76.9}    & \multicolumn{1}{c|}{82.2}    & \multicolumn{1}{c|}{83.8}    & \multicolumn{1}{c|}{80.2}    & \multicolumn{1}{c|}{66.5}    &   \multicolumn{1}{c}{78.7}   \\ \cline{2-14} 
                \multicolumn{1}{c|}{}  & MT3D\cite{ref18}      & \multicolumn{1}{c|}{76.0}  & \multicolumn{1}{c|}{87.6}   & \multicolumn{1}{c|}{89.8}   & \multicolumn{1}{c|}{85.0}   & \multicolumn{1}{c|}{81.2}   & \multicolumn{1}{c|}{75.7}   & \multicolumn{1}{c|}{81.0}    & \multicolumn{1}{c|}{84.5}    & \multicolumn{1}{c|}{85.4}    & \multicolumn{1}{c|}{82.2}    & \multicolumn{1}{c|}{68.1}    &  \multicolumn{1}{c}{81.5}    \\ \cline{2-14} 
                \multicolumn{1}{c|}{}  & 3DLocal\cite{ref24}   & \multicolumn{1}{c|}{78.2}  & \multicolumn{1}{c|}{90.2}   & \multicolumn{1}{c|}{92.0}   & \multicolumn{1}{c|}{87.1}   & \multicolumn{1}{c|}{83.0}   & \multicolumn{1}{c|}{76.8}   & \multicolumn{1}{c|}{83.1}    & \multicolumn{1}{c|}{86.6}    & \multicolumn{1}{c|}{86.8}    & \multicolumn{1}{c|}{84.1}    & \multicolumn{1}{c|}{70.9}    &  \multicolumn{1}{c}{83.7}    \\ \cline{2-14} 
                \multicolumn{1}{c|}{\textbf{CL}}   & GaitGL\cite{ref23}    & \multicolumn{1}{c|}{76.6}  & \multicolumn{1}{c|}{90.0}   & \multicolumn{1}{c|}{90.3}   & \multicolumn{1}{c|}{87.1}   & \multicolumn{1}{c|}{84.5}   & \multicolumn{1}{c|}{79.0}   & \multicolumn{1}{c|}{84.1}    & \multicolumn{1}{c|}{87.0}    & \multicolumn{1}{c|}{87.3}    & \multicolumn{1}{c|}{84.4}    & \multicolumn{1}{c|}{69.5}    &    \multicolumn{1}{c}{83.6}  \\ \cline{2-14} 
                \multicolumn{1}{c|}{}  & CSTL\cite{ref38}      & \multicolumn{1}{c|}{78.1}  & \multicolumn{1}{c|}{89.4}   & \multicolumn{1}{c|}{91.6}   & \multicolumn{1}{c|}{86.6}   & \multicolumn{1}{c|}{82.1}   & \multicolumn{1}{c|}{79.9}   & \multicolumn{1}{c|}{81.8}    & \multicolumn{1}{c|}{86.3}    & \multicolumn{1}{c|}{88.7}    & \multicolumn{1}{c|}{86.6}    & \multicolumn{1}{c|}{\underline{75.3}}    &  \multicolumn{1}{c}{84.2}    \\ \cline{2-14} 
                \multicolumn{1}{c|}{} & TransGait\cite{ref34}   & \multicolumn{1}{c|}{\textbf{80.1}}  & \multicolumn{1}{c|}{89.3}   & \multicolumn{1}{c|}{91.0}   & \multicolumn{1}{c|}{89.1}   & \multicolumn{1}{c|}{84.7}   & \multicolumn{1}{c|}{\textbf{83.3}}   & \multicolumn{1}{c|}{85.6}    & \multicolumn{1}{c|}{87.5}    & \multicolumn{1}{c|}{88.2}    & \multicolumn{1}{c|}{\underline{88.8}}    & \multicolumn{1}{c|}{\textbf{76.6}} & \multicolumn{1}{c}{85.8}    \\ \cline{2-14} 
                \multicolumn{1}{c|}{} & \begin{math}\text{GaitStrip}^{*}\end{math}\cite{ref22} & \multicolumn{1}{c|}{\underline{79.9}}  & \multicolumn{1}{c|}{\textbf{92.3}}   & \multicolumn{1}{c|}{\underline{93.4}}   & \multicolumn{1}{c|}{\underline{89.2}}   & \multicolumn{1}{c|}{\underline{86.0}}   & \multicolumn{1}{c|}{80.0}   & \multicolumn{1}{c|}{\underline{86.0}}    & \multicolumn{1}{c|}{\underline{88.5}}    & \multicolumn{1}{c|}{\underline{91.7}}    & \multicolumn{1}{c|}{87.5}    & \multicolumn{1}{c|}{73.5}    &  \multicolumn{1}{c}{\underline{86.2}}    \\  \cline{2-14} 
                \multicolumn{1}{c|}{} & \textbf{Ours}      & \multicolumn{1}{c|}{73.8}  & \multicolumn{1}{c|}{\underline{91.7}}   & \multicolumn{1}{c|}{\textbf{93.5}}   & \multicolumn{1}{c|}{\textbf{91.5}}   & \multicolumn{1}{c|}{\textbf{87.6}}   & \multicolumn{1}{c|}{\underline{82.6}}   & \multicolumn{1}{c|}{\textbf{87.3}}    & \multicolumn{1}{c|}{\textbf{91.8}}    & \multicolumn{1}{c|}{\textbf{92.9}}    & \multicolumn{1}{c|}{\textbf{88.9}}    & \multicolumn{1}{c|}{72.0}    & \multicolumn{1}{c}{\textbf{\textbf{86.7}}}    \\ 
            \bottomrule[1.5pt]
        \end{tabular}
    }
    \caption*{Superscript \begin{math}\ast \end{math} is for the work\cite{ref22} available on arXiv. Underlining represents the sub-optimal results, while bold represents the best results.}
\end{table}

\begin{table}[H]
    \centering
    \captionsetup{skip=2pt}
    \caption{The overall average Rank-1 accuracy (\%) on CASIA-B.}
    \label{tab:label_2}
    \resizebox{0.45\textwidth}{!}{
        \begin{tabular}{c|c c c|c}
            \toprule[1.2pt]
            Method   & NM & BG & CL & Mean \\ \hline\hline
            GaitSet\cite{ref20}  &  95.0  &  87.2  &  70.4  &   84.2   \\ \hline
            GaitPart\cite{ref5} &  96.2  &  91.5  &  78.7  &   88.8  \\ \hline
            MT3D\cite{ref18}     &  96.7  &  93.0  &  81.5  &   90.4  \\ \hline
            3DLocal\cite{ref24}  &  97.5  &  94.3  &  83.7  &   91.8 \\ \hline
            GaitGL\cite{ref23}   &  97.4  &  94.5  &  83.6  &   91.8  \\ \hline
            CSTL\cite{ref38}     &  97.8  &  93.6  &  84.2  &   91.9   \\ \hline
            TransGait\cite{ref34}  &  \textbf{98.1} & 94.9  &  85.8  &   92.9   \\ \hline
            \begin{math}\text{GaitStrip}^{*}\end{math}\cite{ref22} &  97.6  &  \underline{95.2}  &  \underline{86.2}  &  \underline{93.0} \\ \hline
            Ours     &  \underline{97.9}  &  \textbf{95.8}  &  \textbf{86.7}  &   \textbf{93.5}   \\
            \bottomrule[1.2pt]
        \end{tabular}
    }
\end{table}

\begin{figure}[htb]
    \centering
    \includegraphics[width=1\textwidth]{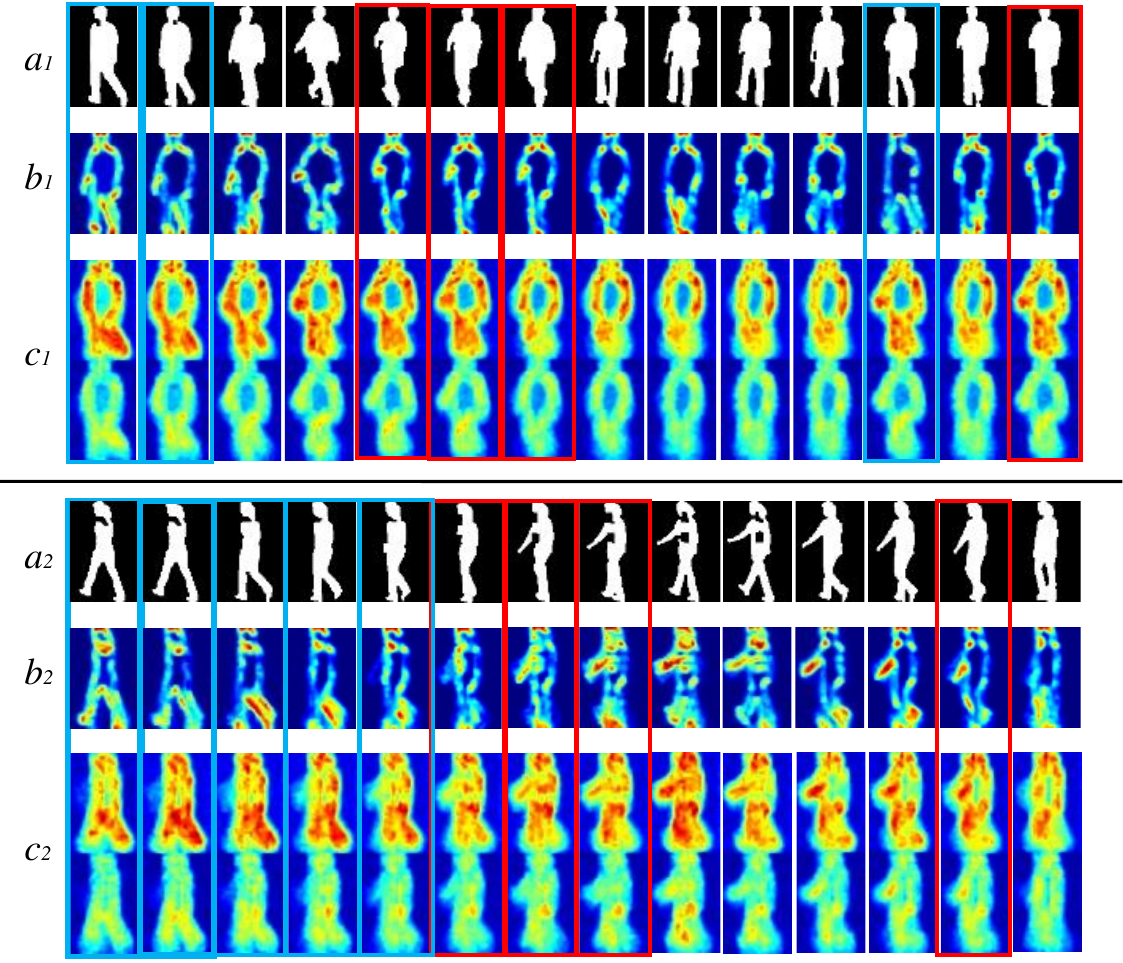}
    \caption{Visualization of the heatmaps for different layers in GaitASMS on CASIA-B. \textbf{(a) Top}: the sequence of the silhouettes; \textbf{(b) Middle}: the heatmaps of the ASRE-1; \textbf{(c) Below}: the heatmaps of the MSTA. The red boxes represent the silhouettes and heatmaps with self-occlusion. The blue boxes represent frames and heatmaps with missing hand contours.}
    \label{fig:heatmap}
\end{figure}

\textbf{Heatmap}. To better explain the effectiveness of our modules, the heatmaps on CASIA-B are shown in Fig.\ref{fig:heatmap}. In \(b_{1, 2}\) and \(c_{1, 2}\), it can be observed that ASRE with the edge-based attention mechanism can adaptively focus on the most discriminative edge dynamic features, especially on the leg and arm parts. As shown in the red and blue boxes in Fig.\ref{fig:heatmap}, even in cases of self-occlusion, limited short-term temporal information and the absence of hand contours, MSTA can still enrich the temporal information and obtain the complete heatmaps through the multi-scale temporal aggregation. For instance, in the first five columns of \(c_{2}\), it can be observed that even in the absence of arms, GaitASMS also effectively supplements the missing parts to address occlusion due to the presence of MSTA. These show that the proposed modules can indeed promote the learning of adaptive structured representations and the multi-scale aggregation of temporal features.

\par \textbf{Performances on OUMVLP}. We also compare our methods with leading video-based approaches, including GEINet, GaitSet, GaitPart, GLN, and GaitGL. Table \ref{tab:label_3} summarizes the performance of these methods on the OU-MVLP dataset. The results indicate that our method outperforms the others, and achieves the highest accuracy in the majority of views, with an average accuracy of 89.9\%. It also shows that our method can be effectively applied to large-scale datasets.

\begin{table}[H]
    \centering
    \captionsetup{skip=2pt}
    \caption{Rank-1 accuracy (\%) on OU-MVLP under 14 probe views, excluding identical identical-view cases.}
    \label{tab:label_3}
    \resizebox{1\textwidth}{!}{
        \begin{tabular}{l|llllllllllllll|l}
            \toprule[1.5pt]
            \multirow{2}{*}{Method} & \multicolumn{14}{c|}{Probe view}                                       & \multirow{2}{*}{Mean} \\ \cline{2-15}
                                    & \multicolumn{1}{c|}{\begin{math}0^{o}\end{math}} & \multicolumn{1}{c|}{\begin{math}15^{o}\end{math}} & \multicolumn{1}{c|}{\begin{math}30^{o}\end{math}} & \multicolumn{1}{c|}{\begin{math}45^{o}\end{math}} & \multicolumn{1}{c|}{\begin{math}60^{o}\end{math}} & \multicolumn{1}{c|}{\begin{math}75^{o}\end{math}} & \multicolumn{1}{c|}{\begin{math}90^{o}\end{math}} & \multicolumn{1}{c|}{\begin{math}180^{o}\end{math}} & \multicolumn{1}{c|}{\begin{math}195^{o}\end{math}} & \multicolumn{1}{c|}{\begin{math}210^{o}\end{math}} & \multicolumn{1}{c|}{\begin{math}225^{o}\end{math}} & \multicolumn{1}{c|}{\begin{math}240^{o}\end{math}} & \multicolumn{1}{c|}{\begin{math}255^{o}\end{math}} & \begin{math}270^{o}\end{math} &                       \\ \hline\hline
            GEINet\cite{ref19}                  & \multicolumn{1}{c|}{11.4}  & \multicolumn{1}{c|}{29.4}   & \multicolumn{1}{c|}{41.5}   & \multicolumn{1}{c|}{45.5}   & \multicolumn{1}{c|}{39.5}   & \multicolumn{1}{c|}{41.8}   & \multicolumn{1}{c|}{38.9}   & \multicolumn{1}{c|}{14.9}    & \multicolumn{1}{c|}{33.1}    & \multicolumn{1}{c|}{43.2}    & \multicolumn{1}{c|}{45.6}    & \multicolumn{1}{c|}{39.4}    & \multicolumn{1}{c|}{40.5}    &    
            \multicolumn{1}{c|}{36.3}    & \multicolumn{1}{c}{35.8}   \\ \hline
            GaitSet\cite{ref20}                 & \multicolumn{1}{c|}{79.5}  & \multicolumn{1}{c|}{87.9}   & \multicolumn{1}{c|}{89.9}   & \multicolumn{1}{c|}{90.2}   & \multicolumn{1}{c|}{88.1}   & \multicolumn{1}{c|}{88.7}   & \multicolumn{1}{c|}{87.8}   & \multicolumn{1}{c|}{81.7}    & \multicolumn{1}{c|}{86.7}    & \multicolumn{1}{c|}{89.0}    & \multicolumn{1}{c|}{89.3}    & \multicolumn{1}{c|}{87.2}    & \multicolumn{1}{c|}{87.8}    &       \multicolumn{1}{c|}{86.2}    & \multicolumn{1}{c}{87.1}   \\ \hline
            GaitPart\cite{ref5}                & \multicolumn{1}{c|}{82.6}  & \multicolumn{1}{c|}{88.9}   & \multicolumn{1}{c|}{90.8}   & \multicolumn{1}{c|}{91.0}   & \multicolumn{1}{c|}{89.7}   &
            \multicolumn{1}{c|}{89.9}   & \multicolumn{1}{c|}{89.5}   & \multicolumn{1}{c|}{85.2}   & \multicolumn{1}{c|}{88.1}    & \multicolumn{1}{c|}{90.0}    & \multicolumn{1}{c|}{90.1}    & \multicolumn{1}{c|}{89.0}    & \multicolumn{1}{c|}{89.1}    & \multicolumn{1}{c|}{88.2}    &       \multicolumn{1}{c}{88.7}    \\ \hline
            GLN\cite{ref21}                  & \multicolumn{1}{c|}{83.8}  & \multicolumn{1}{c|}{90.0}   & \multicolumn{1}{c|}{91.0}   & \multicolumn{1}{c|}{91.2}   & \multicolumn{1}{c|}{90.3}   & \multicolumn{1}{c|}{90.0}   & \multicolumn{1}{c|}{89.4}   & \multicolumn{1}{c|}{85.3}    & \multicolumn{1}{c|}{\textbf{89.1}}    & \multicolumn{1}{c|}{\textbf{90.5}}    & \multicolumn{1}{c|}{\textbf{90.6}}    & \multicolumn{1}{c|}{89.6}    & \multicolumn{1}{c|}{89.3}    &         \multicolumn{1}{c|}{88.5}    & \multicolumn{1}{c}{89.2}     \\ \hline
            GaitGL\cite{ref23}                  & \multicolumn{1}{c|}{84.9}  & \multicolumn{1}{c|}{90.2}   & \multicolumn{1}{c|}{91.1}   & \multicolumn{1}{c|}{91.5}   & \multicolumn{1}{c|}{\textbf{91.1}}   & \multicolumn{1}{c|}{90.8}   & \multicolumn{1}{c|}{90.3}   & \multicolumn{1}{c|}{88.5}    & \multicolumn{1}{c|}{88.6}    & \multicolumn{1}{c|}{90.3}    & \multicolumn{1}{c|}{90.4}    & \multicolumn{1}{c|}{89.6}    & \multicolumn{1}{c|}{\textbf{89.5}}    &         \multicolumn{1}{c|}{88.8}    & \multicolumn{1}{c}{89.7}     \\ \hline
            \textbf{Ours}                    & \multicolumn{1}{c|}{\textbf{85.6}}  & \multicolumn{1}{c|}{\textbf{90.4}}   & \multicolumn{1}{c|}{\textbf{91.2}}   & \multicolumn{1}{c|}{\textbf{91.6}}   & \multicolumn{1}{c|}{\textbf{91.1}}   & \multicolumn{1}{c|}{\textbf{90.9}}   & \multicolumn{1}{c|}{\textbf{90.4}}   & \multicolumn{1}{c|}{\textbf{89.2}}    & \multicolumn{1}{c|}{\textbf{89.1}}    & \multicolumn{1}{c|}{90.4}    & \multicolumn{1}{c|}{90.5}    & \multicolumn{1}{c|}{\textbf{89.8}}    & \multicolumn{1}{c|}{\textbf{89.5}}    &    \multicolumn{1}{c|}{\textbf{89.0}}    & \multicolumn{1}{c}{\textbf{89.9}}  \\ 
            \bottomrule[1.5pt]
        \end{tabular}
    }
\end{table}

\subsection{Ablation Studies}

\par \noindent \textbf{Effectiveness of ASRE.} Unlike most part-based gait recognition methods that only use a fixed segmentation strategy to obtain the parts of human silhouettes, we propose an Adaptive Structured Representation Extraction Module (ASRE), which segments the edges of spatial features in latent space and generates adaptive structured spatial representations. To evaluate the module's effectiveness, several ablation experiments are designed, referred to as Group A. As shown in Table \ref{tab:label_5}, we substituted ASRE with GLConv from GaitGL and MSTA with a standard 3D temporal convolution in A-b, and the average accuracy of A-b is 91.4\%. In contrast, the accuracy of A-e (with ASRE) achieved 91.8\%, resulting in an increase of 0.4\%. Moreover, a comparison between A-d and A-f shows that ASRE improves accuracy by 1.5\%, resulting in a performance of 93.5\%. These results demonstrate the effectiveness of ASRE. The reasons are as follows: \textbf{1)} It utilizes an edge-based attention mechanism to enable the entire model to selectively focus on fine-grained local gait representations, thereby improving the discriminative power of the model; \textbf{2)} It also obtains the global spatial information of gait, which complements the local gait representations and enables our model to learn the key relation between parts.

\begin{table}[H]
    \centering
    \captionsetup{skip=2pt}
    \caption{Ablation Study: Group A. Control condition: w/ and w/o applying ASRM, MSTA, or Random Mask. Results are rank-1 accuracies (\%) on CASIA-B under 11 probe views, excluding identical-view cases.}
    \label{tab:label_5}
        \resizebox{0.75\textwidth}{!}{
           \begin{tabular}{l|lll|lll|l}
                \toprule[1.3pt]
                    \multirow{2}{*}{Group A} & \multicolumn{3}{c|}{Module}   & \multicolumn{3}{c|}{Condition}   & \multicolumn{1}{c}{\multirow{2}{*}{\centering Mean}} \\ \cline{2-7} 
                    & \multicolumn{1}{c}{ASRM} & \multicolumn{1}{c}{MSTA} & \multicolumn{1}{c|}{Random Mask} & \multicolumn{1}{c}{NM} & \multicolumn{1}{c}{BG} & \multicolumn{1}{c|}{CL} &   
                    \\ \hline\hline
                    \multicolumn{1}{c|}{a} & \multicolumn{1}{c}{\begin{math} - \end{math}} & \multicolumn{1}{c}{\begin{math}-\end{math}}   & \multicolumn{1}{c|}{\begin{math} - \end{math}}  & \multicolumn{1}{c}{95.9} & \multicolumn{1}{c}{92.4}  & \multicolumn{1}{c|}{80.4} & \multicolumn{1}{c}{89.6} 
                    \\ \hline
                    \multicolumn{1}{c|}{b} & \multicolumn{1}{c}{\begin{math} - \end{math}} & \multicolumn{1}{c}{\begin{math}-\end{math}}   & \multicolumn{1}{c|}{\begin{math} \surd \end{math}}  & \multicolumn{1}{c}{97.1} & \multicolumn{1}{c}{94.5}  & \multicolumn{1}{c|}{82.7} & \multicolumn{1}{c}{91.4(\(\uparrow\)1.8)} 
                    \\ \hline
                    \multicolumn{1}{c|}{c} & \multicolumn{1}{c}{\begin{math} \surd \end{math}} & \multicolumn{1}{c}{\begin{math} - \end{math}}  & \multicolumn{1}{c|}{\begin{math} \surd \end{math}}  & \multicolumn{1}{c}{97.2}   & \multicolumn{1}{c}{94.6}   & \multicolumn{1}{c|}{83.7}  & \multicolumn{1}{c}{91.8(\(\uparrow\)2.2)}
                    \\ \hline
                    \multicolumn{1}{c|}{d} & \multicolumn{1}{c}{\begin{math} - \end{math}} & \multicolumn{1}{c}{\begin{math} \surd \end{math}}   & \multicolumn{1}{c|}{\begin{math} \surd \end{math}}  & \multicolumn{1}{c}{97.7} & \multicolumn{1}{c}{94.6}  & \multicolumn{1}{c|}{83.7} & \multicolumn{1}{c}{92.0(\(\uparrow\)2.4)} 
                    \\ \hline
                    \multicolumn{1}{c|}{e} & \multicolumn{1}{c}{\begin{math} \surd \end{math}} & \multicolumn{1}{c}{\begin{math} \surd \end{math}}  & \multicolumn{1}{c|}{\begin{math} - \end{math}}  & \multicolumn{1}{c}{97.5}   & \multicolumn{1}{c}{95.2}   & \multicolumn{1}{c|}{\textbf{87.2}}  & \multicolumn{1}{c}{93.3(\(\uparrow\)3.7)} 
                    \\ \hline
                    \multicolumn{1}{c|}{f} & \multicolumn{1}{c}{\begin{math} \surd \end{math}} & \multicolumn{1}{c}{\begin{math} \surd \end{math}}  & \multicolumn{1}{c|}{\begin{math} \surd \end{math}}  & \multicolumn{1}{c}{\textbf{97.9}}   & \multicolumn{1}{c}{\textbf{95.8}}   & \multicolumn{1}{c|}{86.7}  & \multicolumn{1}{c}{\textbf{93.5(\(\uparrow\)3.9)}}
                    \\
                \bottomrule[1.3pt]
            \end{tabular}
        }
\end{table}

\begin{table}[H]
    \centering
    \captionsetup{skip=2pt}
    \caption{Accuracy (\%) of the MSTA with different dilation rates on CASIA-B, under three cloth conditions.} 
    \label{tab:label_6}
    \resizebox{0.4\textwidth}{!}{
        \begin{tabular}{l|lll|l}
            \toprule[1.2pt]
                \multirow{2}{*}{(DCB1, DCB2)} & \multicolumn{3}{c|}{Condition}   & \multirow{2}{*}{Mean} \\ \cline{2-4}  & \multicolumn{1}{c}{NM} & \multicolumn{1}{c}{BG} & \multicolumn{1}{c|}{CL} &   \\ \hline\hline
                \multicolumn{1}{c|}{(1, 2)}   & \multicolumn{1}{c}{97.6}   & \multicolumn{1}{c}{95.1}   & \multicolumn{1}{c|}{86.0}    & \multicolumn{1}{c}{92.9}                       \\ \hline
                \multicolumn{1}{c|}{(2, 4)}   & \multicolumn{1}{c}{\textbf{97.9}}   & \multicolumn{1}{c}{\textbf{95.8}}   & \multicolumn{1}{c|}{\textbf{86.7}}    & \multicolumn{1}{c}{\textbf{93.5}}                       \\ \hline
                \multicolumn{1}{c|}{(4, 8)}   & \multicolumn{1}{c}{97.9}   & \multicolumn{1}{c}{95.3}   & \multicolumn{1}{c|}{86.0}    & \multicolumn{1}{c}{93.1}                       \\
            \bottomrule[1.2pt]
        \end{tabular}
    }
\end{table}

\par \noindent \textbf{Effectiveness of MSTA.} Most of the previous gait recognition methods \cite{ref22, ref23, ref24} only use conventional 3D convolution for short-range temporal modeling. However, due to the subtle differences between adjacent frames \cite{ref28}, it is difficult to extract discriminative temporal features solely through the temporal modeling of adjacent frames. Thus, the MSTA module is proposed, composed of multi-scale dilated convolutional blocks with the residual connection. In Table \ref{tab:label_6}, the combination of dilation rates of 2 and 4 in MSTA can better aggregate multi-scale temporal information, as compared to other combinations of dilation rates. As shown in Table \ref{tab:label_5}, compared to A-b, the addition of MSTA in A-d resulted in an average accuracy improvement of 0.6\%. In the experiments conducted in A-c (with ASRE) and A-f (with ASRE and MSTA), we observed that the average accuracy of the model also increased by 1.7\%. Especially, under BG and CL conditions, the performance improved by 1.2\% and 3\%, respectively. These results demonstrate that the MSTA module is effective in capturing both long-term and short-term temporal information, thereby enhancing the robustness of the overall model to occlusion.\\

\par \noindent \textbf{Effectiveness of the random mask.} Unlike traditional data augmentation methods such as horizontal flipping, rotation and erasing, the random mask applies different masks to randomly selected subjects. In Table \ref{tab:label_4}, the accuracy of random erasing shows a decrease of 2.6\% compared to the baseline. It suggests that the random erasing struggles to simulate long-term self-occlusion effectively and may result in the absence of gait information. In Table \ref{tab:label_7}, from the ablation experiments of mask rates, the accuracy of a mask rate of 0.1 is the highest. Due to the abundant and easily accessible gait information within the NM condition, its impact is minimal at lower mask rates [0.1, 0.3, 0.5]. At mask rates of [0.7, 0.9], a significant loss of valuable gait information hinders the training of the model and the accuracy has significantly decreased. Therefore, all subsequent experiments are conducted with a mask rate of 0.1. As shown in Table \ref{tab:label_5}, the introduction of the random mask improved the accuracy by 1.8\% in the A-a vs. A-b comparison experiments. Especially under the BG and CL conditions, the accuracy improved by 2.1\% and 2.3\%, respectively. This strongly indicates that the random mask can effectively enhance the robustness of the model to long-term self-occlusion.

\begin{table}[H]
    \centering
    \captionsetup{skip=2pt}
    \caption{Accuracy (\%) of different data augmentation on CASIA-B, under three cloth conditions.} 
    \label{tab:label_4}
    \resizebox{0.5\textwidth}{!}{
        \begin{tabular}{l|lll|l}
            \toprule[1.2pt]
                \multirow{2}{*}{Data Augmentation} & \multicolumn{3}{c|}{Condition}   & \multirow{2}{*}{Mean} \\ \cline{2-4}  & \multicolumn{1}{c}{NM} & \multicolumn{1}{c}{BG} & \multicolumn{1}{c|}{CL} &   \\ \hline\hline
                \multicolumn{1}{c|}{w/o Baseline}   & \multicolumn{1}{c}{95.9}   & \multicolumn{1}{c}{92.4}   & \multicolumn{1}{c|}{80.4}    & \multicolumn{1}{c}{89.6}                       \\ \hline
                \multicolumn{1}{c|}{w/ Random Erasing}   & \multicolumn{1}{c}{94.2}   & \multicolumn{1}{c}{89.8}   & \multicolumn{1}{c|}{77.0}    & \multicolumn{1}{c}{87.0}                       \\ \hline
                \multicolumn{1}{c|}{w/ Random Mask}   & \multicolumn{1}{c}{97.1}   & \multicolumn{1}{c}{94.5}   & \multicolumn{1}{c|}{82.7}    & \multicolumn{1}{c}{91.4}                       \\
            \bottomrule[1.2pt]
        \end{tabular}
    }
\end{table}

\begin{table}[H]
    \centering
    \captionsetup{skip=2pt}
    \caption{Accuracy (\%) of the random mask with different mask rates on CASIA-B, under three cloth conditions.} 
    \label{tab:label_7}
    \resizebox{0.4\textwidth}{!}{
        \begin{tabular}{l|lll|l}
            \toprule[1.2pt]
                \multirow{2}{*}{Mask Rate} & \multicolumn{3}{c|}{Condition}   & \multirow{2}{*}{Mean} \\ \cline{2-4}  & \multicolumn{1}{c}{NM} & \multicolumn{1}{c}{BG} & \multicolumn{1}{c|}{CL} &   \\ \hline\hline
                \multicolumn{1}{c|}{0.1}   & \multicolumn{1}{c}{\textbf{97.9}}   & \multicolumn{1}{c}{95.8}   & \multicolumn{1}{c|}{\textbf{86.7}}    & \multicolumn{1}{c}{\textbf{93.5}}                       \\ \hline
                \multicolumn{1}{c|}{0.3}   & \multicolumn{1}{c}{97.8}   & \multicolumn{1}{c}{\textbf{96.1}}   & \multicolumn{1}{c|}{86.5}    & \multicolumn{1}{c}{\textbf{93.5}}                       \\ \hline
                \multicolumn{1}{c|}{0.5}   & \multicolumn{1}{c}{\textbf{97.9}}   & \multicolumn{1}{c}{95.8}   & \multicolumn{1}{c|}{86.4}    & \multicolumn{1}{c}{93.4}                       \\ \hline
                \multicolumn{1}{c|}{0.7}   & \multicolumn{1}{c}{97.7}   & \multicolumn{1}{c}{95.8}   & \multicolumn{1}{c|}{85.3}    & \multicolumn{1}{c}{92.9}                       \\ \hline
                \multicolumn{1}{c|}{0.9}   & \multicolumn{1}{c}{97.3}   & \multicolumn{1}{c}{95.5}   & \multicolumn{1}{c|}{84.9}    & \multicolumn{1}{c}{92.6}                       \\ 
            \bottomrule[1.2pt]
        \end{tabular}
    }
\end{table}

\par \noindent \textbf{Generality of the modules.} To evaluate the generality of proposed modules, we adapt our modules to GaitGL. In Table \ref{tab:label_8}, the baseline GaitGL is denoted as B-a, the accuracy of B-b (with ASRE), B-c (with MSTA), and B-d (with random mask) improved by 0.8\%, 0.3\%, and 0.2\%, respectively. Specifically, In B-b and B-c, the GLFE module and LTA module of GaitGL are respectively replaced by ASRE and MSTA. In B-c, the random mask is introduced into GaitGL. These results significantly validate the portability and effectiveness of the proposed modules.

\begin{table}[H]
    \centering
    \captionsetup{skip=2pt}
    \caption{Ablation Study: Group B. Control condition:  w/ and w/o applying ASRM, MSTA, or Random Mask. This experiment is to verify the generality of ASRM, MSTA, and RM on GaitGL \cite{ref23}.}
    \label{tab:label_8}
    \resizebox{0.75\textwidth}{!}{
        \begin{tabular}{l|lll|lll|l}
            \toprule[1.3pt]
                \multirow{2}{*}{Group B} & \multicolumn{3}{c|}{Module}   & \multicolumn{3}{c|}{Condition}   & \multicolumn{1}{c}{\multirow{2}{*}{\centering Mean}} \\ \cline{2-7} 
                & \multicolumn{1}{c}{ASRM} & \multicolumn{1}{c}{MSTA}  & \multicolumn{1}{c|}{Random Mask} & \multicolumn{1}{c}{NM} & \multicolumn{1}{c}{BG} & \multicolumn{1}{c|}{CL} &   \\ \hline\hline
                \multicolumn{1}{c|}{a}  & \multicolumn{1}{c}{\begin{math}-\end{math}}  & \multicolumn{1}{c}{\begin{math}-\end{math}} & \multicolumn{1}{c|}{\begin{math}-\end{math}}  & \multicolumn{1}{c}{97.4} & \multicolumn{1}{c}{94.5}  & \multicolumn{1}{c|}{83.6} & \multicolumn{1}{c}{91.8} 
                \\ \hline
                \multicolumn{1}{c|}{b} & \multicolumn{1}{c}{\begin{math} \surd \end{math}} & \multicolumn{1}{c}{\begin{math}-\end{math}} & \multicolumn{1}{c|}{\begin{math}-\end{math}}  & \multicolumn{1}{c}{97.6}   & \multicolumn{1}{c}{94.8}   & \multicolumn{1}{c|}{85.5}  & \multicolumn{1}{c}{92.6 (\(\mathbf{\uparrow }\)0.8)}
                \\ \hline
                \multicolumn{1}{c|}{c}  & \multicolumn{1}{c}{\begin{math}-\end{math}}  & \multicolumn{1}{c}{\begin{math} \surd \end{math}} & \multicolumn{1}{c|}{\begin{math}-\end{math}}  & \multicolumn{1}{c}{97.3} & \multicolumn{1}{c}{94.4}  & \multicolumn{1}{c|}{84.6} & \multicolumn{1}{c}{92.1 (\(\uparrow\)0.3)} 
                \\ \hline
                \multicolumn{1}{c|}{d} & \multicolumn{1}{c}{\begin{math}-\end{math}} & \multicolumn{1}{c}{\begin{math}-\end{math}} & \multicolumn{1}{c|}{\begin{math} \surd \end{math}}  & \multicolumn{1}{c}{97.2}   & \multicolumn{1}{c}{94.2}   & \multicolumn{1}{c|}{84.6}  & \multicolumn{1}{c}{92.0 (\(\uparrow\)0.2)}
                \\
            \bottomrule[1.3pt]
        \end{tabular}
    }
\end{table}

\section{Conclusion}
\par \noindent This paper proposes a novel gait recognition framework, denoted GaitASMS, which is based on adaptive structured spatial representation and multi-scale temporal aggregation. By adopting the adaptive edge-based attention mechanism to focus on the most dynamic local edge feature, the GaitASMS can better capture the adaptive structured spatial representations in latent embedding space. Due to the presence of the fundamental weakness caused by the use of short-term temporal window functions, the GaitASMS introduces the multi-scale temporal aggregation module base on dilated 3d convolution to better aggregate the semantic information of gait sequence silhouettes. As a novel data augmentation method, the random mask can enrich the sample space of long-term occlusion and enhance the generalization of the model. Experiments on two well-known public gait datasets, CASIA-B and OU-MVLP, have indicated that compared with other SOTA gait recognition methods, GaitASMS achieves the highest accuracy, especially in occlusion conditions. We hope this work brings the attention of researchers to the adaptive extraction of dynamic local spatial features in gait, as well as the method of utilizing dilated convolutions to achieve both long and short-term temporal aggregation.

\section*{Acknowledgments}
\noindent This work is funded by the National Natural Science Foundation of China, grant number: 62002215; This work is funded by Shanghai Pujiang Program (No. 20PJ1404400).

\section*{Declarations}
\noindent \textbf{Conflict of interest} All authors declare that they have no conflicts of interest.
\par \noindent \textbf{Data availability} The datasets that support the findings of this study are openly available in 10.1109/ICIP.2011.6115889, reference number \cite{ref32, ref37}.

\bibliographystyle{ieeetr}
\bibliography{bibfile}
\end{document}